\documentclass[twoside,leqno,twocolumn]{article}
\usepackage{graphicx} % Required for inserting images
\usepackage{xcolor}
\usepackage{graphicx}
\usepackage{algorithmic}
\usepackage{algorithm}
\usepackage{amsmath}
\usepackage{graphics}
\usepackage{tabularx}
\usepackage{ltexpprt}
\usepackage{subcaption}

\newcommand\blfootnote[1]{%
  \begingroup
  \renewcommand\thefootnote{}\footnote{#1}%
  \addtocounter{footnote}{-1}%
  \endgroup
}

\begin{document}
\title{Identification and Uses of Deep Learning Backbones via Pattern Mining}
\author{Michael Livanos\thanks{University of California Davis, USA. \texttt{mjlivanos@ucdavis.edu}}
\and Ian Davidson\thanks{University of California Davis, USA. \texttt{davidson@cs.ucdavis.edu}}}

\date{}

\maketitle

\fancyfoot[R]{\scriptsize{Copyright \textcopyright\ 2024 by SIAM\\
Unauthorized reproduction of this article is prohibited}}
%%
%% The abstract is a short backbone of the work to be presented in the
%% article.
\begin{abstract}
Deep learning is extensively used in many areas of data mining as a black-box method with impressive results. However, understanding the core mechanism of how deep learning makes predictions is a relatively understudied problem. Here we explore the notion of identifying a backbone of deep learning for a given group of instances. A group here can be instances of the same class or even misclassified instances of the same class. We view each instance for a given group as activating a subset of neurons and attempt to find a subgraph of neurons associated with a given concept/group. We formulate this problem as a set cover style problem and show it is intractable and presents a highly constrained integer linear programming (ILP) formulation. As an alternative, we explore a coverage-based heuristic approach related to pattern mining, and show it converges to a Pareto equilibrium point of the ILP formulation. Experimentally we explore these backbones to identify mistakes and improve performance, explanation, and visualization. We demonstrate application-based results using several challenging data sets, including Bird Audio Detection (BAD) Challenge and Labeled Faces in the Wild (LFW), as well as the classic MNIST data.
\end{abstract}


\section{Introduction}

As models are deployed to tasks traditionally only trusted to humans, understanding a model's behavior is often required. This is particularly true for methods such as deep learning (DL), as their decision-making mechanisms are inherently opaque. Existing work in explainable artificial intelligence (XAI) provides interpretability by explaining a prediction decision on a single instance. While this provides some insight into a particular prediction, it does not demystify the more general decision-making process of the learner. Further, such explanation mechanisms also suffer from an overreliance on the input space. Such explanations are convenient for interpretable input spaces, such as images or text, but may be useless for data with uninterpretable feature spaces, such as embeddings or audio spectrograms as we study in Section \ref{sec:exp}.

In this paper, we explore the area of creating backbones of a deep learner. These backbones can be used for a variety of tasks including identifying mistakes, improving prediction, and global explanation.

\begin{table}[]
\caption{Taxonomy of XAI into three categories, each with distinct goals and definitions of interpretability.}
\label{tab:taxonomy}
\resizebox{\columnwidth}{!}{%
\begin{tabular}{lll}
\hline
\textbf{Category}                                                                               & \begin{tabular}[c]{@{}l@{}}Interpretability\\ Defintion\end{tabular}                                                                                                  & Examples                                                                                       \\ \hline
\multicolumn{1}{l|}{\begin{tabular}[c]{@{}l@{}}Explaining\\ Model\\ Decisions\end{tabular}}     & \multicolumn{1}{l|}{\begin{tabular}[c]{@{}l@{}}Justify a model's\\ action on a\\ particular instance\end{tabular}}                                                    & \begin{tabular}[c]{@{}l@{}}LIME {[}16{]}\\ Counterfactual\\ Explanations {[}22{]}\end{tabular} \\ \hline
\multicolumn{1}{l|}{\begin{tabular}[c]{@{}l@{}}Creating\\ Interpretable\\ Models\end{tabular}}  & \multicolumn{1}{l|}{\begin{tabular}[c]{@{}l@{}}Create an inherently\\ interpretable model\\ OR distill an opaque\\ model into an \\ interpretable model\end{tabular}} & \begin{tabular}[c]{@{}l@{}}Distilling Networks\\ into Decision Trees\\ {[}6{]}\end{tabular}    \\ \hline
\multicolumn{1}{l|}{\begin{tabular}[c]{@{}l@{}}Investigating\\ Model\\ Mechanisms\end{tabular}} & \multicolumn{1}{l|}{\begin{tabular}[c]{@{}l@{}}Provide deeper\\ understanding of\\ how the model\\ processes instance\end{tabular}}                                   & \begin{tabular}[c]{@{}l@{}}Feature\\ Visualization {[}13{]}\\ Ours\end{tabular}               
\end{tabular}
}
\end{table}

\smallskip
\noindent
\textbf{Core Idea.}
A core insight is that any instance activates a subset of neurons in the network. Hence, a concept backbone is a subgraph of hidden units that frequently co-activate for a subset of instances associated with a concept such as a group of instances of a class incorrectly predicted, or some other phenomenon we wish to explain. We can find a collection of concept backbones which are for different concepts and are \underline{distinct/different} from each other. We refer to this as a collective backbone.

For example, given a network meant to distinguish dogs from cats, one can find a concept backbone for the concept of \underline{mispredicting} dogs as cats in order to identify future mispredictions and another for correctly predicting dogs to better understand the model's decision-making process. Further, exploiting the distinction between these two provides a basis for accomplishing more complex tasks such as correcting mistakes (Section \ref{sec:expD}).

Our approach is flexible enough to answer a variety of questions. We demonstrate our work on three domains: Labeled Faces in the Wild (LFW) \cite{lfw}, the Bird Audio Detection Challenge (BAD) \cite{stowell2019automatic}, and MNIST for visual explanations. In the more challenging LFW dataset, we demonstrate the robustness of our method even when faced with small data, 12-way classification, and class imbalance. We show the versatility of our method by also showing its utility on the non-image datasets of the BAD challenge. Our backbones have high coverage distinctness (the subgraph to cover minimally covers instances from other categories).

We make the following contributions with the last point being important, as justifying the utility of an explanation is critical.
\begin{itemize}
    \item We present the backbone identification problem for supervised prediction as a coverage problem (See Problem Definition and Formulation), formulate it as an ILP, and prove intractability (See Theorem 1).
    \item We provide a heuristic algorithm to find a completely connected subgraph covering many instances for the novel concept of a concept-level backbone. A collection of these backbones can explain an entire model (See Section \ref{sec:App}).
    \item We prove that this algorithm will produce a solution that is Pareto optimal to the problem in respect to the maximizing problem objective and minimizing relaxation. (See Theorem 2)
    \item We explore sixteen different networks in three domains (See Section \ref{sec:exp}). Specifically:
    \begin{itemize}
        \item We apply feature visualization to create explanations from our backbones for the MNIST dataset.
        \item We use our backbones to \underline{identify} mispredictions with high confidence in the LFW and BAD datasets.
        \item We use multiple backbones to \underline{correct} those mispredictions to a great deal of success in the BAD networks.
    \end{itemize} 
\end{itemize}

The paper is organized as follows: we first discuss the core problem and show its intractability, after which we describe our approach. We design and complete our experiments next and then conclude.

\blfootnote{ In the interest of space, theorems, proofs, and algorithm 2 are provided at: www.cs.ucdavis.edu/~davidson/BackboneThms.pdf}

\section{Overview of Our Approach}
\label{sec:App}
\smallskip
\noindent
\textbf{Backbone Desiderata.}
The output of our approach is a concise subgraph of the deep learner that activates with a particular concept/group. We begin by defining the characteristics of a good collective backbone and then a high-level problem formulation. 

\smallskip
\noindent
Every concept-level backbone must:
\begin{enumerate}
    \item Describe \underline{all} the members of the concept
    \item Be \underline{distinct} from all opposing concepts
    \item Be \underline{concise} in terms of size
\end{enumerate}

To exemplify our reasoning, consider the following explanations of
%the concept of
dogs. One explanation may be ``A domesticated four-legged animal with a tail". This is unsatisfactory as this could describe other animals. One can tailor this response to exclude other animals by including details such as ``a long snout", however this may exclude some dogs such as pugs or bulldogs. Finally, consider the response by the Oxford Languages Dictionary: ``A domesticated carnivorous mammal that typically has a long snout, an acute sense of smell, nonretractable claws, and a barking, howling, or whining voice"\cite{mw}. While this is specific and inclusive, in certain contexts, the eloquence of a smaller explanation may be desirable. In the same sense, backbones must be descriptive, exclusive, and ideally concise.

Through this example, one can see that these criteria are at odds with each other. To make an explanation general enough to apply to all members, one may need to sacrifice conciseness. To ensure that the explanation is distinct from other related concepts, one may need to make generalizations that exclude members of the group. This is why backbone discovery naturally lends itself to an optimization setup.

\smallskip
\noindent
%\textbf{Explanation vs backbone.}
%The output of our approach is a concise subgraph of the deep learner that activates with a particular concept/group. We can interpret this as being the relevant part of the learner associated with the concept which can be used in a variety of ways from visualizing the typical that activates it (see Figure \ref{fig:virtual}) to identifying and correcting mispredictions (see Section \ref{sec:expD}).
%Later, we visualize these subgraphs to provide perceptive incredibility, and therefore differentiate these two presentations with the terms backbone and explanation respectively.

\begin{figure}
    \centering
    \includegraphics[scale = 0.22]{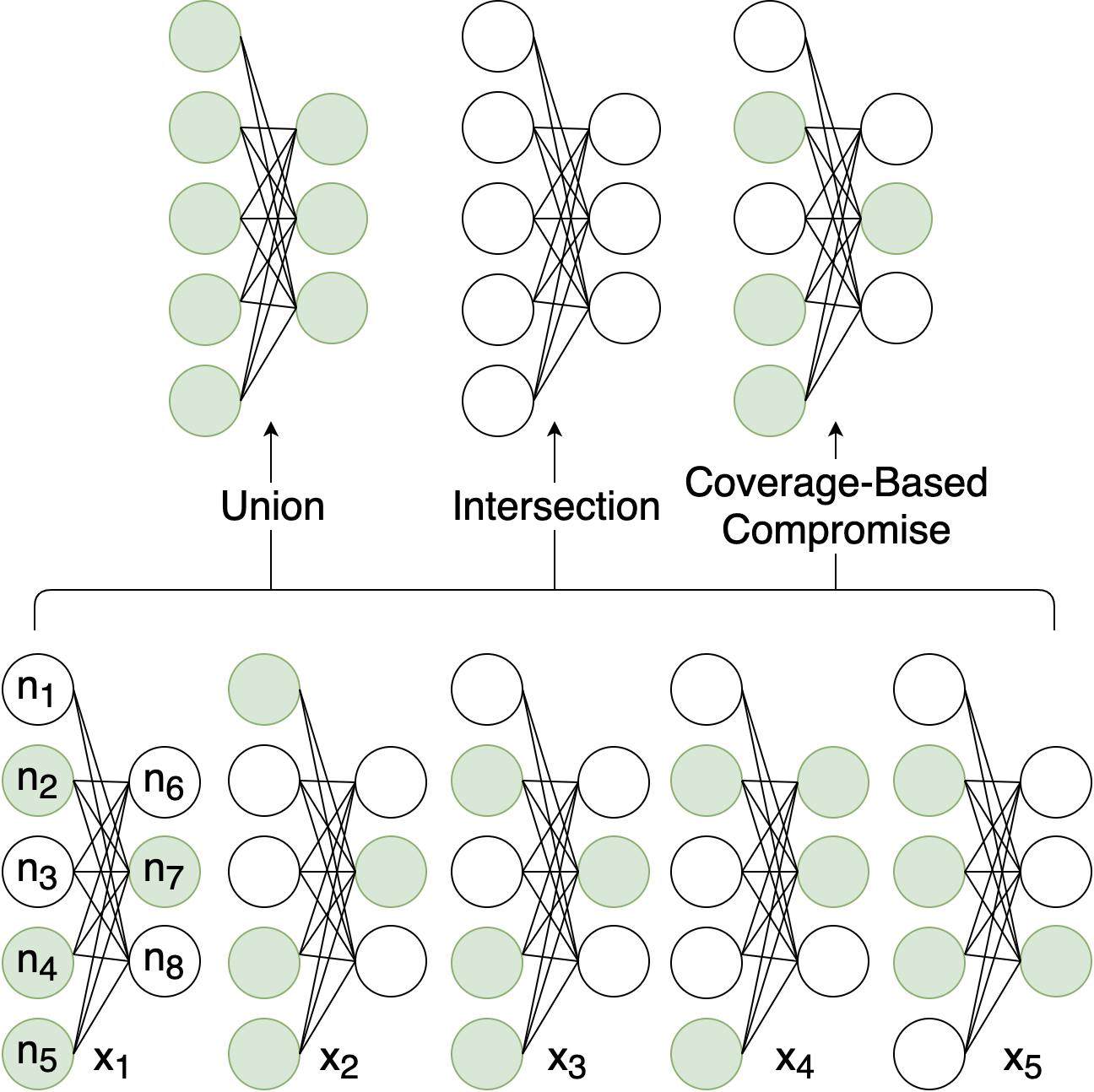}
    \caption{Potential issues with taking the union and intersection of activation vectors. In this example, there are eight neurons in the network and five instances in the concept. Neurons $n_2, n_4, n_5,$ and $n_8$ form the clearest summaries, occurring in 80\% of the instances and the other neurons in only 20\%. The intersection is empty since it requires neurons to be present in all instances, and the union is the whole network since it requires neurons to be present in only once.}
    \label{fig:union_int}
\end{figure}

\noindent
\textbf{Concept-Level backbone as Finding a Minimal Graph.}  
We view each instance $x_k$ as activating a subset $N_k$ of the model's hidden layer neurons by creating an activation vector of each's nodes influence with each component corresponding to a hidden neuron in the network. \emph{Influence} is calculated as the absolute value of the neuron's activation times the sum of the absolute value of weights. 
We create a set of node activations $C_i = \{N_1 \dots N_m\}$ for the $i^{th}$ concept for all $m$ instances associated with this concept. 
Naturally, these vectors can be viewed as graphs in the network or transactions. We then investigate $C_1 \ldots C_n$ to understand which neurons are quintessential for the classification of concept $i$. 
The goal of a CL-backbone is to find a subgraph of the network such that the nodes in the subgraph are connected and cover many instances in the concept.
%We can visualize a backbone as shown in Figure \ref{fig:expExample}.
This is a challenging problem as the naive approach of taking the union or intersection of the transactions for a given concept yields only trivial backbones (see Figure \ref{fig:union_int}). 

Below we formalize the problem for a single backbone, and later expand it to a collection of backbones:

\smallskip
\noindent
\textbf{The Concept Summarization Problem.}
Given a set of graphs $G_1 \ldots G_n$ of the DL node influences for $n$ instances, find a backbone (subgraph) $G^*$ such that $\forall i$:
\begin{itemize}
    \item \textbf{Complete Coverage} $G^* \subset G_i $
    \item \textbf{Connected} $G^*$ is a connected graph and 
    \item \textbf{Conciseness} $|G^*|$ is minimal.
\end{itemize}

The extension to the Collective backbone problem is then to find multiple CL-summaries ($G^*_1 \ldots G^*_k$) with the additional requirement of \textbf{Distinctness} from each other ($G^*_1 \cap G^*_2 \cap \ldots \cap G^*_k = \emptyset $ for $k$ concepts).
This problem can be easily translated to an ILP, however we prove this to be intractable in Theorem 1, and in practice often infeasible to satisfy. Even relaxations of this ILP are extremely computationally expensive. Instead, we design a coverage-based approach to efficiently find such subgraphs which we prove produces an optimal result to the original formulation of this problem.

\iffalse
\begin{figure}
    \centering
    \includegraphics[scale = 0.35]{exampleExplanation.png}
    \textcolor{red}{This can't be followed from just looking at it.} 
    \caption{Model level backbone composed of four concept level summaries (color) generated for a Bird Audio Detection Challenge network, with opacity denoting importance of each subgraph (discussed in section \ref{sec:heuristic}). Neurons not used in the backbone are collapsed to enhance readability.}
    \label{fig:expExample}
\end{figure}
\fi

\section{Problem Definition and ILP Formulation}
\label{sec:formalizing}

Let $X$ be a set of $n$ data points which, for clarity, are all instances in the same concept (e.g. all incorrectly predicted instances of the same class).  Let $M$ be a learned DL model consisting of fully connected nodes $R$ where $R_{l,j}$ is the $j^{th}$ hidden node at layer $l$. Further, let $W$ be the weights in the DL with $W_{n_1,n_2}$ being the weight connecting node $n_1$ to $n_2$. There is no need for $M$ to be trained from $X$, but this is the case in our experiments. Further, let $x_k$ be the $k^{th}$ instance of the data which activates the subset of nodes $N_k$, that is $N_k \subset R$. We describe the requirements for activation later. Then $X$ has an analogous representation $N = \{N_1 \ldots N_n\}$ which is a set of subsets of node activations that can be represented as a binary $n \times |R|$ table, $T$, with the entry $T_{k,j}=1$ representing that instance $k$ activates node $j$. This is naturally a transaction dataset with the items being node activation. The threshold associated with node activation is application-specific and in our experiments we threshold so that the $r$ most influential neurons are used. We lift the above notation to any number of categories by using the superscript $i$.

\smallskip
\noindent
\textbf{Definitions.} Before discussing the ILP using notation, we provide definitions for some of the concepts we hope to achieve. In the strict ILP, the backbone must cover all of the instances. That is, the backbone must reflect the activation of all instances $x_i \in X$. This is referred to as \textbf{complete coverage}. To create a collective backbone, we also enforce \textbf{orthogonality}, in that no two backbones should share a common neuron. In the relaxed ILP, these constraints are no longer strict, and instead, we use the terms \textbf{coverage} and \textbf{diversity} to refer to the idea that the backbone should cover most, but not necessarily all, instances, and that there should be minimal overlap between backbones.

\noindent
\textbf{Problem Statements.} 
Our problem is to find a subset of nodes $R^i_*$ which explains all instances in concept $i$.  
A naive version of the problem is to find the largest (hence most descriptive) backbone, which will be equivalent to taking the union of $N^i$ overall $i$, that is, $R^i_* = \cup_j N_j^i$.  However, this risks creating a huge network, and likely produces high overlap between CL-Explanations. Similarly, the intersection over $N^i$ ($R^i_* = \cap_j N_j^i$) is likely to yield a very small subset of nodes unlikely to form a connected subgraph. 

Instead, we model this discovery problem as a set cover style problem. First, we describe the formulation for a single-concept backbone and then extend it to the collective backbone by adding an orthogonality requirement.

\smallskip
\iffalse
\textbf{The Concept-Level backbone  Problem}
\begin{eqnarray}
argmin_{R_*}~~~\sum_i |R^i_*| s.t. \\ \nonumber
~~R^i_*  \subset N^i_j ~~\forall i,j~~\texttt{Complete Coverage} \\ \nonumber
~~R^i_* \cap R^j_* = \emptyset~~\forall i,j,~~i \neq j~~\texttt{Orthogonality}
\end{eqnarray}
\fi
\begin{align*}
argmin&_{R_*}~~~\sum_i |R^i_*| &\\ 
s.t.&~~~R^i_*  \subset N^i_j ~~\forall i,j &&\hspace{-0.33in} \texttt{Complete Coverage} \\ 
s.t.&~~~R^i_* \cap R^j_* = \emptyset~~\forall i,j,~~i \neq j &&\texttt{Orthogonality}\\
\end{align*}

However, this will return a subset of nodes ($R^i_*$) for each class $i$, each of which may not define a connected sub-network, that is, they may not contain a node at each layer in the DL, and there need not even be any connections between the nodes (i.e., non-zero weights). Hence we require any sub-network we find as possessing two properties: i) \textit{Layer Inclusion:} There is at least one node in $R^*$ for each layer in the original network and ii) \textit{Connectivity:} A path using non-zero edge weights between every node in $R^i_*$ exists. For simplicity, we define $C_{R^i_*}(j,k)$ as being the multiplication of the absolute value of the weights that connect nodes $j$ and $k$ using only nodes in $R^i_*$.  Hence the problem we attempt to solve in this paper is given below:

\smallskip
\noindent
\textbf{The Connected Concept-Level Backbone Problem.}
\begin{align*}
argmin&_{R_*} \sum_i |R^i_*| & \\ 
s.t. &~~~R^i_*  \subset N^i_j ~~\forall i,j &&\hspace{-0.4in}\texttt{Complete Coverage} \\ 
s.t. &~~~R^i_* \cap R^j_* = \emptyset~~\forall i,j,~~i \neq j &&\hspace{-0.07in}\texttt{Orthogonality}  \\
s.t. &~~~\exists  n \in R^i_*: n \in R_j~~\forall j~~\forall i &&\hspace{-0.24in}\texttt{Layer~Inclusion} \\
s.t. &~~~\exists C_{R^i_*}(j,k) > 0~~\forall j,k \in R^i_*~~\forall i &&\texttt{Connectivity}
\end{align*}

A proof for intractability of this problem is provided in Theorem 1. Replacing the first two constraints with relaxations may be a solution: not all instances need be covered/explained, but instead, $\delta_i$ can be forgotten for concept $i$, and its description has up to $\gamma_i$ overlapping nodes with other descriptions.

\smallskip
\noindent
\textbf{The Relaxed Connected Concept-Level Backbone Problem.}
\begin{align*}
arg&min_{R_*} \sum_i |R^i_*|  & \\ 
s.t. &~~~(N^i_j - R^i_*) \geq \delta^i~~\forall i,j && \hspace{0.09in}\texttt{ Coverage} \\ 
s.t. &~~~(N^i_j - R^i_* * R^j_*) \geq \gamma^i ~~\forall i,j,~~i \neq j && \texttt{ Diversity} \\  
s.t. &~~~\sum \delta < p_1  && \hspace{-0.69in}\texttt{ Coverage Relaxation} \\
s.t. &~~~\sum \gamma < p_2  && \hspace{-0.62in} \texttt{ Support Relaxation} \\
s.t. &~~~\exists  n \in R^i_*: n \in R_j~~\forall j~~\forall i &&\hspace{-0.3in}\texttt{Layer~Inclusion} \\
s.t. &~~~\exists C_{R^i_*}(j,k) > 0~~\forall j,k \in R^i_*~~\forall i &&\hspace{-0.1in} \texttt{Connectivity}
\end{align*}

Where $p_1$ and $p_2$ are the maximum number of instances that can be forgotten and the number of overlapping nodes, respectively.

\section{Approach}
\label{sec:heuristic}

In this section, we discuss our heuristic-based solution to the \emph{Connected Concept-Level backbone Problem}. In later sections, we mathematically prove and empirically demonstrate that this algorithm is guaranteed to provide an optimal result. We accomplish this through two simple but efficient algorithms: Find Max Minsup (FMM) (Algorithm \ref{alg:fmm}), which finds the connected layer-inclusive subgraph with the highest support, and F-Score Thresholding (Algorithm 2), which iteratively adds new neurons with the greatest support that either form or adds to a complete graph, to the backbone which maximizes our heuristic.

FMM is a frequent subgraph mining algorithm that finds the most frequent subgraph that meets the fully connected and inclusive layer constraints. Frequent subgraph mining requires a minimum support threshold to be specified \cite{apriori} \cite{fpg}, however there is no way to immediately know what value of minsup will generate a frequent subgraph satisfying the constraints of the problem. To deal with this, minsup is first set to 100\% of the data (the intersection of the transactions) and decremented by a single instance each iteration until a complete (layer inclusive and connected) graph is created, returning the value of support of that subgraph. While it may sound appealing to perform binary search to find this value, frequent pattern mining algorithms tend to grow exponentially in computational complexity as minsup becomes smaller \cite{fpgComplex}, so the method presented is more efficient. Furthermore, since minsup can be decremented to zero, and since each transaction has at least one connected neuron from each layer, we are guaranteed to find the single subgraph with the greatest support. We dub this subgraph the backbone.

\begin{figure}[!h]
    \centering
    \includegraphics[scale = 0.43]{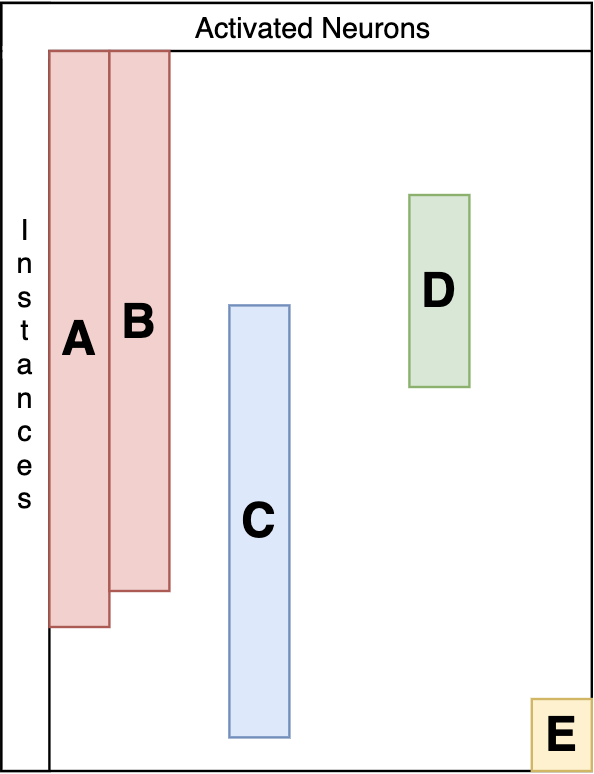}
    \caption{A visualization of the matrix of node activations $N$ as a series of transactions with columns as different neurons and rows as instances. Color corresponds to patterns, and groups of neurons are labeled. FMM only finds group A, but ignores everything else. F-Score thresholding allows groups B and C to be included in the backbone despite having lower support than max minsup. Groups D and E have much lower support, so they will not be included.}
    \label{fig:fscore}
\end{figure}

The backbone only includes the most frequent complete graph, however, and will ignore patterns that are nearly as frequent (see Figure \ref{fig:fscore}). F-Score Thresholding finds a Pareto optimal solution regarding the ILP's objective and minimizing the relaxations. That is, the graph returned cannot simultaneously have greater coverage, diversity, and/or be smaller (see Theorem 2).

We accomplish this by viewing the backbone as a predictive model for which neurons will appear in a transaction and iteratively adding the next most frequent pattern until the change in F-score after one iteration is negative. To calculate an F-Score, we define the true positives as the simultaneous occurrence of a neuron in the backbone and transaction, a false positive as a neuron occurring in the backbone, but not the transaction, and a false negative as a neuron occurring in a transaction contains a neuron. Precision and recall are determined in traditional ways, and F-Score is the harmonic mean between the two. It is important to recognize that maximizing recall maximizes the instances covered, whereas precision acts as a check, penalizing the heuristic for adding infrequent neurons. In order to distinguish between the patterns, a weight is given to each pattern in the backbone equal to that pattern's support divided by max minsup.

\iffalse
\begin{figure}[h!]
\label{fig:approach}
\centering
\includegraphics[scale = 0.25]{Exp_pipeline.png}
\caption{Model pipeline from the input (instances and network) to the output, a weighted backbone (weight denoted by opacity).}
\end{figure}
\fi

\begin{algorithm}[H]\
    \caption{Input: set of activation vectors $N$\newline
    Output: Value of minsup that prodcues a graph with the highest support}
    \begin{algorithmic}
        \STATE $s \leftarrow 1; \ Minimum \ Support$
        \STATE $d \leftarrow 1/len(N); Support \ decrement$
        \STATE $subgraph \leftarrow patternMining(N,s)$
        \WHILE{$\neg completeGraph(subgraph)$}
        \STATE $s = s - d$
        \STATE $subgraph = patternMining(N,s)$
        \ENDWHILE
        \RETURN $s$
    \end{algorithmic}
    \label{alg:fmm}
\end{algorithm}
\hfill

\section{Models and Datasets}
\label{datasets}

In order to demonstrate our technique's invariance to domain and utility on very different types of networks, we conducted experiments using sixteen different networks in three different domains, raw image data from the MNIST dataset, audio data from the Bird Audio Detection Challenge, and embeddings generated from Facenet of faces from the Labeled Faces in the Wild (LFW) dataset. The Section \ref{supData} provides a description of the datasets, model architectures, and why each of these networks is interesting. Each backbone is referred to by the dataset used to create them, with a superscript + indicates that the backbone for correctly predicted instances of a given class and - for incorrectly predicted instances. Unless + or - is given, it is assumed to be the + backbone.\footnote{Trained models, datasets, results, and intermediary results are included at https://github.com/MLivanos/backbonesSDM24}

\section{Experimental Design}
\label{sec:expD}

These results are based on summaries generated from all folds for each network. Details on the datasets, networks, and cross-fold validation are provided in Section \ref{supData}.

\begin{figure*}[h]
    \centering
    \includegraphics[scale = 0.45]{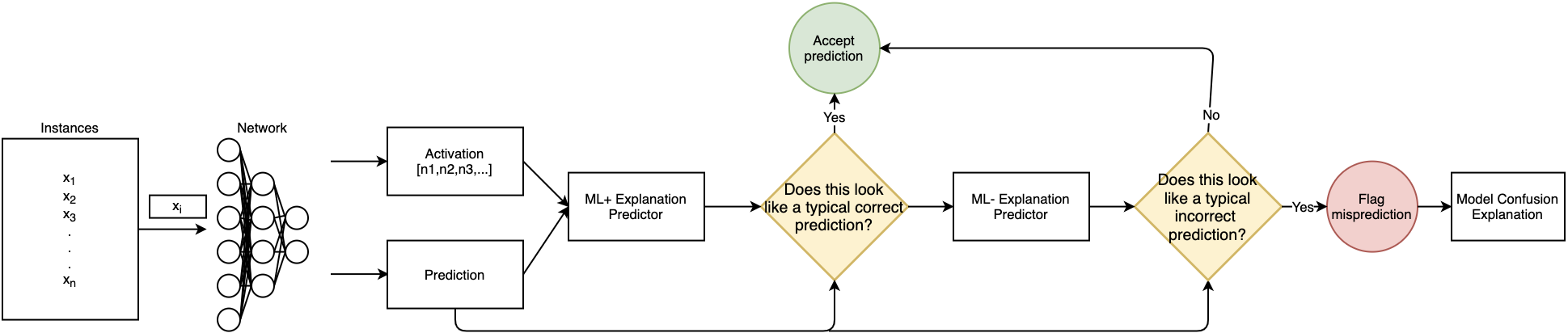}
    \caption{Flow diagram of the process of flagging mispredictions and correcting them using the collective backbone and the prediction of the network.}
    \label{fig:predictingMispredictions}
\end{figure*}

\smallskip
\noindent
\textbf{Heuristic Approach Compared to ILP.}
Before demonstrating interpretability, we compare the heuristic-based solution to that of the ILP. To compare the solutions of the two approaches, we see how two metrics, coverage and overlap, change as new subgraphs are added to the backbone and use relaxed formulations of the ILP as baselines. We say that an instance $x$ is covered by a backbone $C_i$ if some complete connected subgraph exists in the activation vector of $x$ that also exists in $C_i$. Overlap between two or more summaries is the number of neurons that both have in common with each other divided by the size of the summaries. In the case of the BAD network, the ILP is relaxed until solutions can be generated in 24 hours or less, and in the LFW network, they are relaxed until Pareto optimality. Ideally, we would see coverage monotonically increase and diversity minimally decrease as new patterns are added until it reaches a desirable point on the Pareto front.

\smallskip
\noindent
\textbf{Backbone as a Predictive Device.} If a backbone of a concept (such as a class) is robust enough, one should identify the concept when they encounter it. Here, we compare the activation vector for a given instance to each of the CL-summaries, and assign a prediction to the most similar one. Since these are summaries of the concept, we do not expect the accuracy of the backbone to be as high as that of the model, however we do expect a good backbone to have comparable results.

\smallskip
\noindent
\textbf{Predicting Mispredictions and Correcting Them.} We create a pipeline for detecting and correcting mispredictions of the model, summarized in Figure \ref{fig:predictingMispredictions}. As opposed to the previous experiment, we consider both the activation vector for a given instance and the network's prediction of that instance. Using the same method described in the above experiment, we compare the activation vector to the set of correctly predicted CL models, asking the question: "Does this look like a typical correct prediction of the predicted class?" If the prediction of the network and that of the summaries differ, we repeat the process on the incorrectly predicted instances, asking the question "Does this look like a typical misprediction of this class?". If the answer to the first question is no and the second yes, the prediction is assumed a misprediction.

After being flagged, an alternative prediction is provided. For binary classification, this is trivial, as the prediction is simply swapped to the other class. In the case of multi-class classification, a model confusion backbone is provided, a CL backbone in which the concept is "class $x$ being predicted as class $y$" for all combinations of mispredictions, and this backbone is used to determine the alternative prediction.

\smallskip
\noindent
\textbf{Subgraph Visualization.} We perform feature visualization to create human interpretable explanations to find virtual instances in the input space that most activate the subgraph returned from our method. We use particle swarm optimization (PSO) \cite{pso} to minimize the euclidean distance between the normalized activation vector and the backbone returned from our method. The result is a virtual instance whose activation is high in the neurons of the backbone but not any other neurons. Because it would be difficult to leverage the cost function gradient to optimize activation for neurons on different layers, PSO was chosen as the algorithm for this task since it is a gradient-free optimization algorithm.

In order to focus the algorithm and produce crisper explanations, we create a pixel whitelist of the top 40 percent of pixels found in each class. While this reduces and focuses the input space, the optimization stays the same.

\section{Experiments}
\label{sec:exp}

\smallskip
\noindent
\textbf{Heuristic Approach vs ILP.} Theorem 2 proves that our algorithm finds a non-trivial Pareto-optimal solution with respect to minimizing the objective of the problem and the two relaxations, however it is not immediately discernible where on the Pareto-front the solution will lie. Further, while the ILP formulation of the problem is proven intractable in Theorem 1, we empirically examine the speedup of the heuristic solution. In this experiment, we create 15 explanations via our approach and compare them to that of an ILP to demonstrate the viability and efficiency of our algorithm.

\begin{figure}
     \centering
     \begin{subfigure}[b]{0.45\textwidth}
         \centering
         \includegraphics[width=\textwidth]{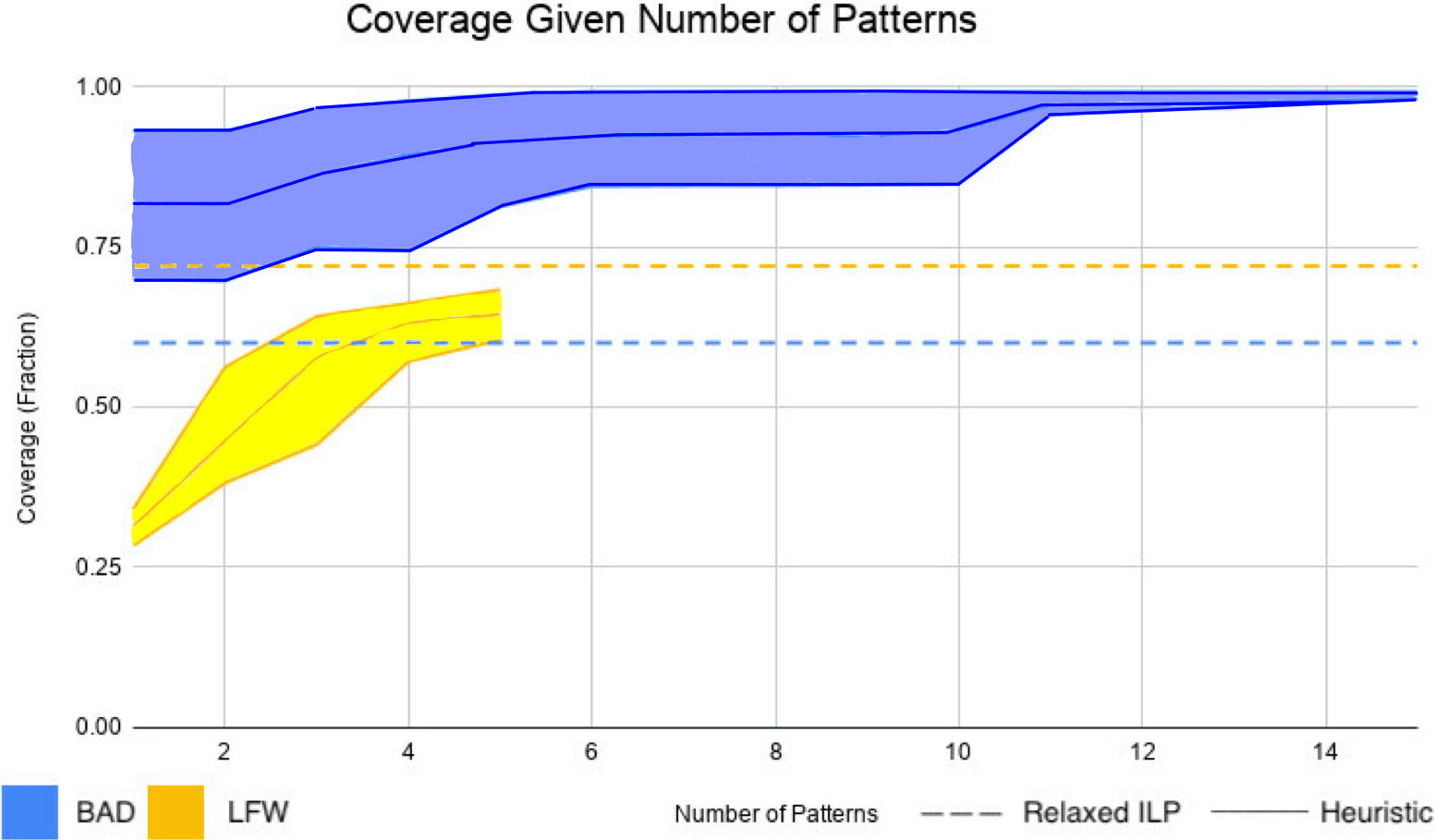}
         \caption{Coverage (vertical axis) against number of patterns added (x-axis).(higher is better)}
         \label{fig:covComp}
     \end{subfigure}
     \hfill
     \begin{subfigure}[b]{0.45\textwidth}
         \centering
         \includegraphics[width=\textwidth]{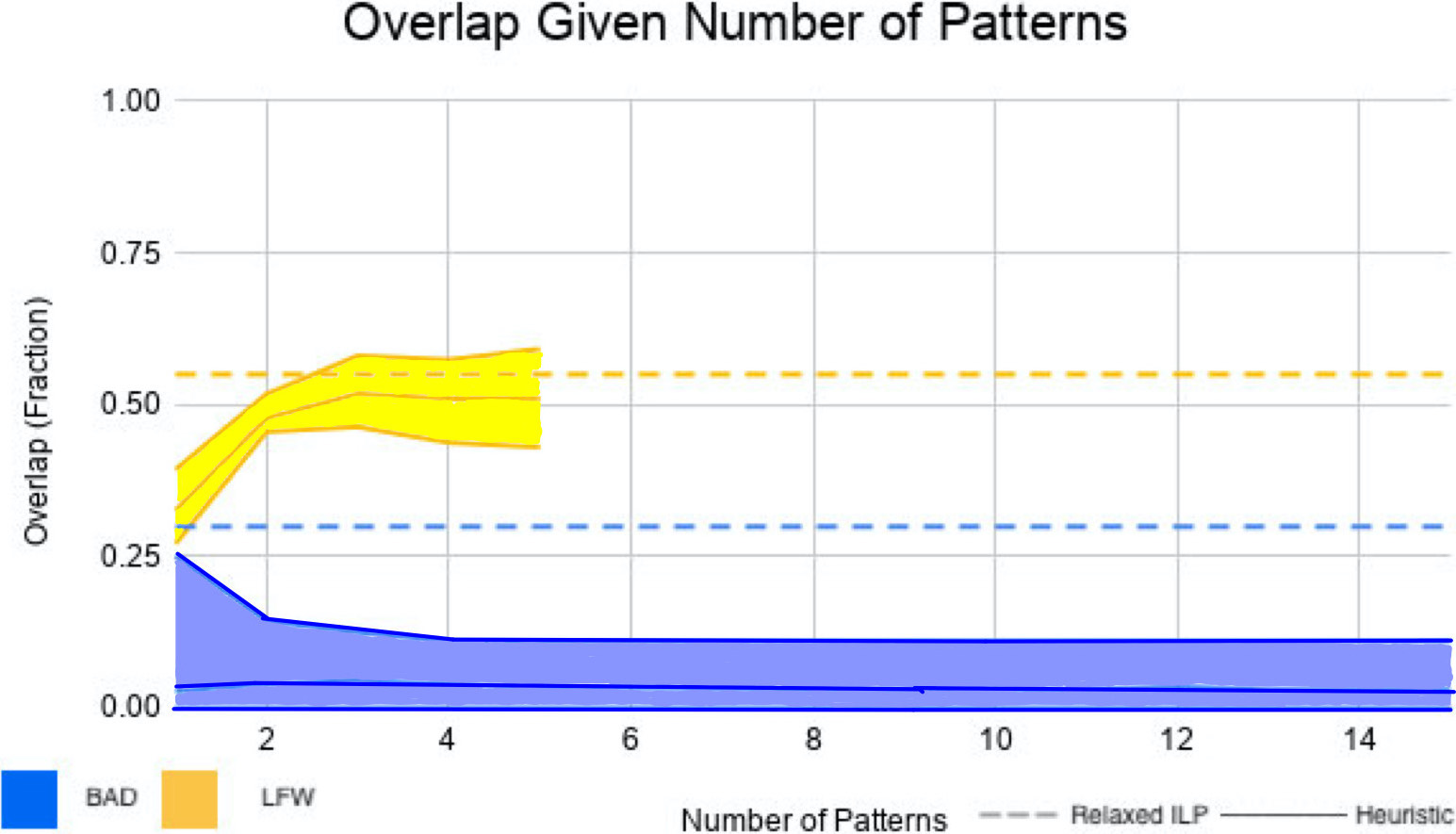}
         \caption{Overlap (vertical axis) against number of patterns added (x-axis). (lower is better)}
         \label{fig:sepComp}
     \end{subfigure}
     \caption{Quantifying coverage and overlap difference between the relaxed ILP and heuristic. For both datasets, the top line represents the maximum (across folds) for that metric, the middle the median, and the lower the minimum. Coverage increases over iterations while overlap minimally increases.}
     \hfill
     \label{fig:comps}
\end{figure}

In the BAD network, the ILP must be relaxed greatly in order to return a solution within 24 hours. As shown in Figure \ref{fig:comps}, the heuristic solution achieves an initial 80\% coverage, and after adding 14 additional patterns, achieves nearly 100\%, compared to the ILP's 60\%. Overlap starts at about 2.8\% and only increases to 3\% at the end, compared to the ILP's 30\%. Not only does our method substantially outperform the relaxed ILP on both metrics, it does so in an average of 12 minutes compared to the ILP's 24 hours, speeding up the process by a factor of 120.

The LFW summaries also drastically increase in coverage as patterns are added to the backbone. In this case, we see that the solution returned by the heuristic approach sacrifices some accuracy for diversity. Five patterns are added before the algorithm terminates; increasing coverage increases from 31.1\% to 64.5\%, compared to the ILP's 72\%. While there is a jump in overlap from the first pattern added to the second, afterward overlap does not significantly increase, ending at 51\% compared to the ILPs 55\%. Moreover, it takes 10 minutes, compared to the ILP's required 30 minutes.

\begin{figure}
    \centering
    \includegraphics[scale = 0.34]{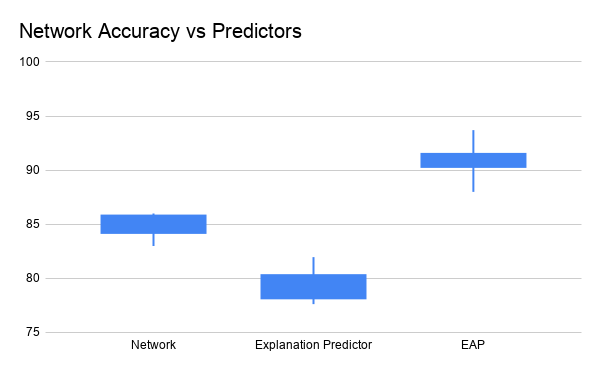}
    \caption{The network, backbone as a predictive model, and the explanation augmented predictor accuracy on BAD test data. When used as a predictive device, the backbone underperforms the network, as expected, however when one considers both the backbone and the output of the network, as one does in the EAP, accuracy is increased significantly.}
    \label{fig:networkVEAP}
\end{figure}

\smallskip
\noindent
\textbf{Predictive Device.} We demonstrate the quality of our explanations by showing that they alone can be used for classification and achieve similar results to that of the networks they explain. If the explanation of the model's behavior can be used in this way, then the explanation is good and faithful to the model.

In both the LFW and the BAD summaries, predictive capabilities had lower but comparable accuracy than their respective models. The median accuracy for the LFW summaries was 45\% compared to the median model's 60\%, and the median accuracy for BAD summaries was 78.5\% compared to the model's 85.2\%. Interestingly, despite having lower accuracy, the predictive device can correctly classify instances that the model could not. This observation was the impetus for creating the explanation augmented predictor.

\smallskip
\noindent
\textbf{Explanation Augmented Prediction.} To demonstrate the practical utility of our explanations, we use them to identify when the models tend to fail and correct their predictions.

In the BAD Challenge dataset, we correctly flag nearly two-thirds of incorrectly predicted instances while only incorrectly flagging 5\% of correctly predicted instances as mispredictions. This allowed us to create a model of greater predictive power than the original network, elevating the median accuracy from 85.2\% to 91.3\%. In addition, all ten folds exhibited an increase in accuracy ranging from 5 to 7.3\%. See Figure \ref{fig:networkVEAP}.

In the LFW dataset, we correctly identify 21.4\% of incorrectly predicted instances, and incorrectly flag 12.1\% of correctly predicted instances. While 33\% of mispredictions could be corrected using the model confusion explanation, augmenting these networks would, on average, have lower accuracy than the model on its own. While the LFW explanations can identify when errors occur, it cannot reliably correct them likely due to the small data nature of the problem.

\begin{figure}
    \begin{subfigure}[b]{0.45\textwidth}
        \centering
        \includegraphics[scale = 0.12]{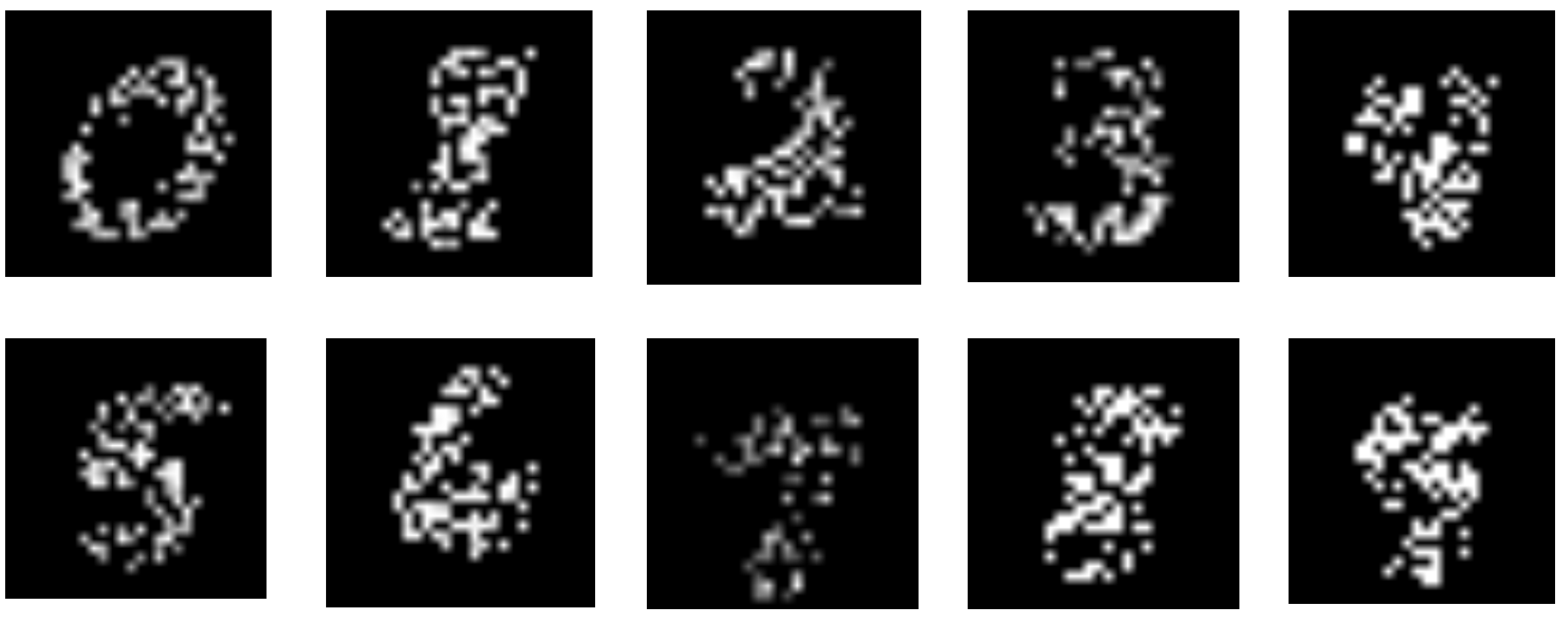}
        \caption{Visualization of the semantic meaning for each concept (digit) in the collective for the MNIST dataset, accomplished via activation maximization)}
        \label{fig:virtual}
        \hfill
    \end{subfigure}
    \begin{subfigure}[b]{0.45\textwidth}
        \centering
        \includegraphics[scale = 0.08]{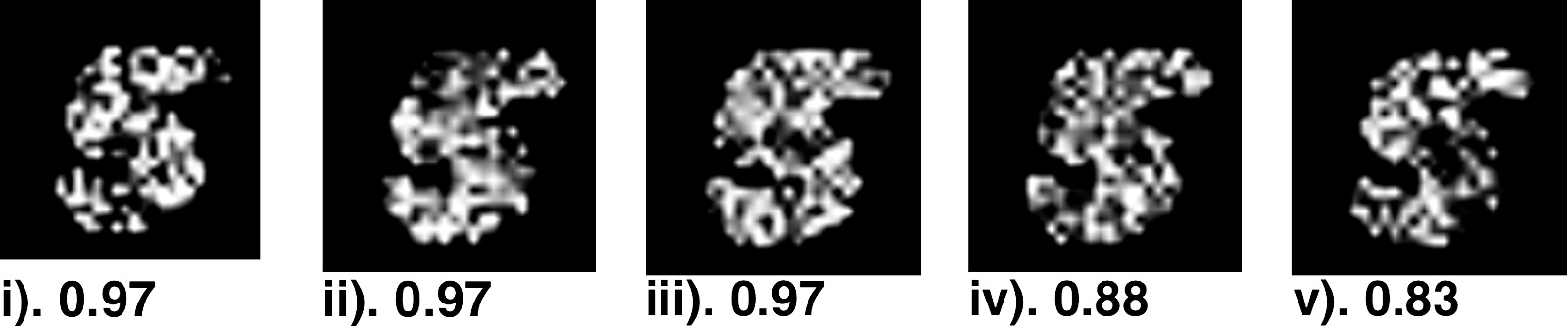}
        \caption{Visualization of the semantic meaning of each component subgraph of the CLE of the digit 5. Different parts of the digit are activated by different subgraphs. For example, ii. focuses on the tail of the 5, whereas v. focuses on the head, and i. on the middle section. Under each instance is the support that subgraph has over the instances.}
        \label{fig:virtual2}
        \hfill
    \end{subfigure}
\end{figure}

\smallskip
\noindent
\textbf{Subgraph Visualization.} The neuron visualization technique provides, in human interpretable terms, the semantic meaning of the graphs returned in Algorithms \ref{alg:fmm} \& \ref{alg:Fscore}. The results of this approach are in Figure \ref{fig:virtual}. This provides insight and validity to the backbone, allowing the user to understand how it is activated. Figure \ref{fig:virtual2} shows the maximization of different subgraphs in the backbone. This allows the user to see how the component subgraphs capture different parts of the concept of each digit, such as the tail vs. the head of the five.

\section{Reproducibility Details: Model Architecture and Dataset Selection}
\label{supData}

\textbf{FaceNet Embeddings From Labeled Faces In The Wild.}
FaceNet was created by Schroff et al 2015 to generate embeddings that have small Euclidian distance between two tight-cropped faces of the same person, but have larger Euclidian distance between different people\cite{schroff2015facenet}. Scroff et al show that embeddings have smaller distances comparing the same face from different angles and lighting than a different face in the same angle and lighting.

We use a subset of the Labeled Faces In The Wild dataset to create a single 12-way classification network trained on the embeddings for each individual. Those individuals were those who has at least 50 images in the dataset. The dataset has significant class imbalance, with the least represented individuals (Serena Williams and Jacques Chirac) having only 53 images, and the most well-represented person (George W. Bush) having over 500. This network is composed of five fully connected hidden layers with 80, 60, 40, 30, and 20 neurons respectively. We train five networks using cross-fold validation with a class-balanced test set and median accuracy of 60\% for the 12-way classification task.

Since our approach requires at least one neuron per layer in the backbone, the network's wide design means that we will have long, spanning summaries. Further, the class imbalance in training, low network accuracy, and lack of available data will pose challenges that our technique will need to overcome. These challenges make explaining this network  the hardest task.

\textbf{Bulbul: Bird Audio Detection Challenge.} Bulbul was developed by Grill et al 2017 as part of the Bird Audio Detection (BAD) challenge. Mel spectrograms generated from raw WAV files are given as input, most of them 10 seconds long. The training data comes from multiple sources, each from different regions of the world, different recording equipment, and different class balance.\cite{stowell2019automatic} Bulbul was the winning network of the challenge in 2018, achieving an area under the curve (AUC) of 88.7\%\cite{grill2017two}. Here, we recreate bulbul using 10-fold cross validation and achieve a median accuracy of 84\% on the validation set. Two notable difference between the networks that we trained and that of Grill et al is that, at the time of the experiments, The Machine Listening Lab (the entity in charge of the BAD Challenge) has not published the labels of their testing data, so we use a subset of the training dataset (that was not used for training of our network) as testing data. Also, our model was trained on a fixed number of epochs, while Grill et al use a variable training scheme. We also report our findings in terms of accuracy, rather than AUC as Grill et al did, and demonstrate that, using backbone to augment prediction, we can significantly improve accuracy. Due to these differences, we do not claim superiority over Grill et al's method, but we do demonstrate that our method can surpass a similar network with the same architecture and training data on the metric of accuracy.

Bulbul is a much shallower network, with only two dense layers with 256 and 32 neurons respectively. This will create smaller summaries than the LFW summaries. Further,  this network is for binary classification and is trained on large datasets (over 10,000 training instances). Due to these factors, we suspect that summaries for this network will work well with our experiments and yield us better results. Finally, this network was chosen to showcase our technique's result on a network of high domain importance, as this was the top model of the 2018 BAD challenge.

\textbf{MNIST} The MNIST digit recognition network is trained directly from the MNIST training set. It is composed of two convolutional layers, with 32 and 64 channels, respectively, each with a 3x3 sliding window. Following each of the convolutional layers is a max pooling layer with a 2x2 sliding window, connected to two fully connected layers of size 64 and 32 neurons.

\section{Related Work}

In the preceding sections, we have discussed our approach and demonstrated its utility on complex XAI tasks. Here, we discuss other approaches and some of their deficiencies which our approach has overcome. As opposed to our work, which provides category and model-level explanations grounded in model architecture, most existing XAI methods explain a model's behavior on specific instances and/or ground their explanation in input space. In this section, we highlight the need for our particular form of explanation in contrast to existing methods.

Many XAI methods provide local interpretability, that is, an explanation for a single instance\cite{Das2020}. Popular techniques that provide this incredibility include those that isolate superpixels of an instance \cite{ribeiro2016should}, \cite{IG}, or counterfactual explanations\cite{white2019measurable}  which generate virtual instances to explain why a different action was not taken. This provides limited insight to the future behavior of the model because it only speaks for the instance that it is explaining and with no guarantee that the model will behave the same way for future instances.

The input space is a natural choice for presenting an explanation, as it is often inherently interpretable, facilitating human perceptive explanation \cite{xaiSurvey}. However, numerous works towards in adversarial attacks demonstrate how engineered, imperceptible changes are drastically alter a network's prediction\cite{szegedy2014intriguing}, \cite{Su_2019}. Grounding an explanation in the architecture, by contrast, focuses exclusively on how the network processes information, allowing a more complete examination of the model's behavior. One example of this deficiency exists in the field of feature visualization. This model-level explanation technique grounds the semantic meaning of layers or neurons in the input space \cite{olah2018the}. Researchers note that semantically different images can achieve similar levels of activation \cite{Olah}, demonstrating the volatile nature of input space as the basis for explanation. 

\section{Conclusion}
\label{sec:conclusion}
We formulate the problem of discovering concept and collective backbones as subgraphs of a deep learner, prove that the ILP formulation of this problem is intractable and expensive even with relaxations. We propose a heuristic-based approach via frequent subgraph mining techniques and prove that this method returns a Pareto optimal result with respect to maximizing the problem's objectives and minimizing relaxations and does so at a fraction of the runtime.

These summaries provide a basis for completing complex XAI tasks that existing methods cannot. Our approach can determine patterns in model failure which can be used to determine mispredictions, patterns in model success, and use the combination of those two to correct mispredictions.
For example, our method succeeded in boosting the performance of the BAD network and could successfully identify errors in the LFW network, although it could not correct them. This indicates that our method performs best on high-performance and binary classification networks trained; however, further investigation is required to understand which of these factors is most pressing. 

This work differs from most XAI research as it presents model level summaries grounded in hidden-neuron space that can be used in ways to improve the trustworthiness of a model and provide greater insight into how the learner makes decisions. 

\noindent
\textbf{Acknowledgments:}~ We thank the reviewers 
for providing helpful suggestions.
This work was supported in part by NSF Grant IIS-1910306 titled:
``Explaining Unsupervised Learning: Combinatorial Optimization Formulations,
Methods and Applications''.

\bibliographystyle{splncs04}
\bibliography{ref}

\begin{thebibliography}{10}
\providecommand{\url}[1]{\texttt{#1}}
\providecommand{\urlprefix}{URL }
\providecommand{\doi}[1]{https://doi.org/#1}

\bibitem{apriori}
Agrawal, R., Srikant, R., et~al.: Fast algorithms for mining association rules.
  In: Proc. 20th int. conf. very large data bases, VLDB. vol.~1215, pp.
  487--499 (1994)

\bibitem{Das2020}
Das, A., Rad, P.: Opportunities and challenges in explainable artificial
  intelligence (xai): A survey (2020)

\bibitem{mw}
Dictionary, O.E.: Dog

\bibitem{fpg}
Gosta~Grahne, J.Z.: Fast algorithms for frequent itemset mining using fp-trees.
  IEEE Transactions On Knowledge And Data Engineering  \textbf{17} (2005)

\bibitem{grill2017two}
Grill, T., Schl{\"u}ter, J.: Two convolutional neural networks for bird
  detection in audio signals. In: 2017 25th European Signal Processing
  Conference (EUSIPCO). pp. 1764--1768. IEEE (2017)

\bibitem{lfw}
Huang, G.B., Mattar, M., Berg, T., Learned-Miller, E.: Labeled faces in the
  wild. workshop on faces in ’real-life’ images. FFINRIA  (2008)

\bibitem{pso}
Kennedy, J., Eberhart, R.: Particle swarm optimization. In: Proceedings of
  ICNN'95. vol.~4, pp. 1942--1948 vol.4 (1995)

\bibitem{fpgComplex}
Kosters, W.A., Pijls, W., Popova, V.: Complexity analysis of depth first and
  fp-growth implementations of apriori. In: International Workshop on Machine
  Learning and Data Mining in Pattern Recognition. pp. 284--292. Springer
  (2003)

\bibitem{Olah}
Olah, C., Mordvintsev, A., Schubert, L.: Feature visualization. Distill  (2017)

\bibitem{olah2018the}
Olah, C., Satyanarayan, A., Johnson, I., Carter, S., Schubert, L., Ye, K.,
  Mordvintsev, A.: The building blocks of interpretability. Distill  (2018).
  \doi{10.23915/distill.00010}, https://distill.pub/2018/building-blocks

\bibitem{ribeiro2016should}
Ribeiro, M.T., Singh, S., Guestrin, C.: " why should i trust you?" explaining
  the predictions of any classifier. In: Proceedings of the 22nd ACM SIGKDD
  international conference on knowledge discovery and data mining. pp.
  1135--1144 (2016)

\bibitem{schroff2015facenet}
Schroff, F., Kalenichenko, D., Philbin, J.: Facenet: A unified embedding for
  face recognition and clustering. In: Proceedings of the IEEE conference on
  computer vision and pattern recognition. pp. 815--823 (2015)

\bibitem{stowell2019automatic}
Stowell, D., Wood, M.D., Pamu{\l}a, H., Stylianou, Y., Glotin, H.: The first
  bird audio detection challenge. Methods in Ecology and Evolution
  \textbf{10}(3),  368--380 (2019)

\bibitem{Su_2019}
Su, J., Vargas, D.V., Sakurai, K.: One pixel attack for fooling deep neural
  networks. IEEE Transactions on Evolutionary Computation  \textbf{23}(5),
  828–841 (Oct 2019). \doi{10.1109/tevc.2019.2890858}

\bibitem{IG}
Sundararajan, M., Taly, A., Yan, Q.: Axiomatic attribution for deep networks
  (2017)

\bibitem{szegedy2014intriguing}
Szegedy, C., Zaremba, W., Sutskever, I., Bruna, J., Erhan, D., Goodfellow, I.,
  Fergus, R.: Intriguing properties of neural networks (2014)

\bibitem{xaiSurvey}
Tjoa, E., Guan, C.: A survey on explainable artificial intelligence {(XAI):}
  towards medical {XAI}. CoRR  \textbf{abs/1907.07374} (2019)

\bibitem{white2019measurable}
White, A., d'Avila Garcez, A.: Measurable counterfactual local explanations for
  any classifier  (2019)

\end{thebibliography}

\end{document}